\documentclass[
hf
]{ceurart}

\usepackage{listings}
\usepackage{subcaption}
\usepackage{tcolorbox}
\usepackage{float}
\usepackage{placeins}
\usepackage{xcolor}

\begin{document}

\copyrightyear{}
\copyrightclause{}
\conference{}



\title{Peeking Inside LLMs: Leveraging Internal Artifacts of LLMs for Enhancing Reliability in Legal Classification}
\tnotemark[1]
\tnotetext[1]{This work has been accepted at the International Workshop on \textit{Automated Semantic Analysis of Information in Law} (\textbf{ASAIL} 2026), co-located with the \textit{International Conference on Artificial Intelligence and Law} (ICAIL 2026).}




\author[1]{Sudipta Santra}[%
email = sudiptasantra232@gmail.com,
]
\fnmark[1]

\author[1]{Debtanu Datta}[%
email = debtanudatta04@gmail.com
]
\fnmark[1]

\author[1]{Saptarshi Ghosh}[%
email = saptarshi@cse.iitkgp.ac.in
]

\address[1]{
Indian Institute of Technology Kharagpur, Kharagpur, 721302, West Bengal, India}

\fntext[1]{These authors contributed equally.}

%
\begin{abstract}
Large Language Models (LLMs) are increasingly being adopted in the legal domain. However, despite their strong performance, LLMs are prone to generating incorrect or hallucinated outputs, raising serious concerns about their reliability in high-stakes domains such as law. 
Detecting the correctness of responses of LLM-based systems is therefore a critical challenge.
In this work, we explore the potential of leveraging internal artifacts of LLM to detect the correctness of their predictions in legal-domain classification tasks. 
We develop approaches that utilize features derived from these internal artifacts to build downstream classifiers capable of identifying incorrect LLM outputs. We evaluate our approach on two representative legal classification tasks: \textit{bail decision prediction} and \textit{statute violation prediction}. 
Our experimental results demonstrate that LLMs' internal artifacts are reliable indicators for detecting incorrect predictions in legal classification tasks, and can be applied to enhance the reliability of LLM-based classification systems.
\end{abstract}


\makeatletter
\def\@copyrightLine{}
\makeatother

\maketitle

\section{Introduction}


Large Language Models (LLMs) have recently demonstrated remarkable performance across a wide spectrum of ML / NLP tasks, including classification, machine translation, summarization, and question answering~\cite{brown2020language,chowdhery2023palm}. 
Naturally, LLMs are being increasingly adopted in the legal domain, where numerous research works highlight their effectiveness in handling challenging tasks such as statute identification, judgment prediction, and more~\cite{aletras2016predicting,datta2026advantages,deroy2025investigating,nigam2025nyayaanumana}.

Despite these advances, \textbf{LLMs are known to generate, at times, incorrect or hallucinated responses}~\cite{huang2025survey, abdullahi2026rise}. 
This limitation is particularly concerning in high-stakes domains such as law, where erroneous predictions can lead to significant legal and ethical consequences. 
Ensuring the reliability and trustworthiness in LLM-generated outputs is therefore crucial in the legal domain.
Recent research has revealed that the internal artifacts of LLMs -- such as self-attention values, hidden fully-connected vectors, token-level probabilities, etc. -- encode informative signals about the model's predictions and uncertainty~\cite{snyder2023early, datta2026lowresource, ji2024internal}. 
These insights suggest that LLMs inherently possess latent information that can be utilized to assess the correctness of their outputs. 
However, despite this growing body of work, \textit{leveraging these internal artifacts / signals for correctness detection in legal-domain tasks remains largely underexplored}. 

To address this gap, we investigate whether the internal artifacts of LLMs can be systematically leveraged to assess the correctness of their outputs in legal classification tasks. 
We develop a Correctness Detection (CD) module that utilizes the internal artifacts derived from LLMs' internal representations, and checks the correctness of a response generated by a LLM.
We develop various \textbf{LLM+CD systems} that are more reliable than the LLMs alone. 
We evaluate the effectiveness of our approach on two popular binary classification tasks in the legal domain: (i)~\textit{bail decision prediction} and (ii)~\textit{statute violation prediction}.
Through comprehensive experiments, we demonstrate how LLMs' internal signals can serve as an additional layer of verification, thereby enhancing the trustworthiness of LLM-based legal decision-support systems. 
To summarize, our contributions are as follows:
\begin{itemize}
    \item We address the task of detecting correctness of LLM responses for legal classification tasks. 
    We develop approaches that leverages LLMs' internal artifacts to build correctness detectors (CD).
    \item We evaluate our approaches across three LLMs on two popular classification tasks in the legal domain: \textit{bail decision prediction} and \textit{statute violation prediction}.
    \item Our experiments demonstrate that internal artifacts provide strong signals of correctness of LLM predictions. Our CDs achieve more than 85\% AUROC scores in detecting correctness of LLM responses in most cases.
    \item Furthermore, we demonstrate that \textbf{our approaches can be used to not only detect correctness, but also enhance the reliability of legal classification}. 
    For instance, for the bail decision prediction task, while the Llama-8b LLM has an accuracy of 54.84\% by itself, one of our LLM+CD systems enhances the accuracy to 74.85\% by reversing the LLM prediction if needed. 
    The accuracy is further enhanced to 84.52\% when the LLM+CD system is given the option of 
    refraining from answering if there is uncertainty about the LLM response.
\end{itemize}
The implementations of the proposed methods are publicly available at: \url{https://github.com/iamDebtanu/LLM_Correctness_Detection_Legal}.

\section{Related Work}
\label{sec:related}


\paragraph{Classification Tasks in the Legal domain.} 
Classification tasks 
arise across multiple settings in the legal domain, including bail decision prediction, statute violation prediction, question answering, and so on. 
In recent times, pretrained LLMs have demonstrated strong performance across a variety of legal classification tasks~\cite{jiang2023legal_syllogism_prompting,dai2025enhancing, sivakumar2025predictive}. 
Prior works have proposed prompting strategies based on legal syllogism to improve reasoning in legal judgment prediction~\cite{jiang2023legal_syllogism_prompting}, while other approaches focus on enhancing LLM performance through task-specific prompting and adaptation techniques~\cite{dai2025enhancing}. 
However, these works consistently observed that LLMs often produce incorrect or unreliable outputs, particularly in complex legal settings where precise reasoning is critical. 
Motivated by these observations, we aim to develop methods to detect the correctness of LLM outputs and improve the reliability of their predictions.

Several benchmark datasets have been introduced to address classification tasks in the legal domain, each targeting different prediction settings. 
For example, CAIL2018~\cite{xiao2018cail2018} focuses on tasks such as charge prediction and the classification of relevant articles. 
SwissJudgment~\cite{niklaus2021swiss} is a benchmark for legal outcome prediction, supporting both binary and multi-class classification tasks. The ECHR~\cite{Chalkidis1} dataset addresses statute violation prediction by focusing on whether specific articles of the European Convention on Human Rights have been violated.
In the Indian context, 
the ILDC dataset~\cite{Vijit1} is designed for bail prediction. 
We use the ECHR and ILDC datasets for experiments in this work.\\


\noindent \textbf{Correctness Detection over LLM Responses. }
In view of incorrect responses and hallucinations of LLMs, there has been recent focus on 
detecting correct / incorrect responses generated by LLMs. 
Some popular approaches for detecting incorrect responses / hallucinations are \textit{internal artifact-based methods} that leverage internal parameters of transformer-based architectures, such as logits, hidden states, and attention matrices, to identify incorrect generations~\cite{snyder2023early, datta2026lowresource, ji2024internal}. 
On the other hand, sampling-based methods rely on LLM-generated sampled responses rather than examining model internal artifacts~\cite{farquhar2024semantic, manakul2023selfcheckgpt}.\\ 

\noindent \textbf{Novelty of our work.}
\textit{Although internal artifact-based approaches have shown effectiveness across NLP tasks, their applicability to legal domain-specific tasks remains limited}. To address this gap, this work develops approaches based on LLMs' internal artifacts for checking the correctness of LLM responses in classification tasks in the legal domain. 







\section{Background on Internal Artifacts of LLMs}
\label{sec:background}

\begin{figure}[t]
\centering
\includegraphics[width=1\linewidth]{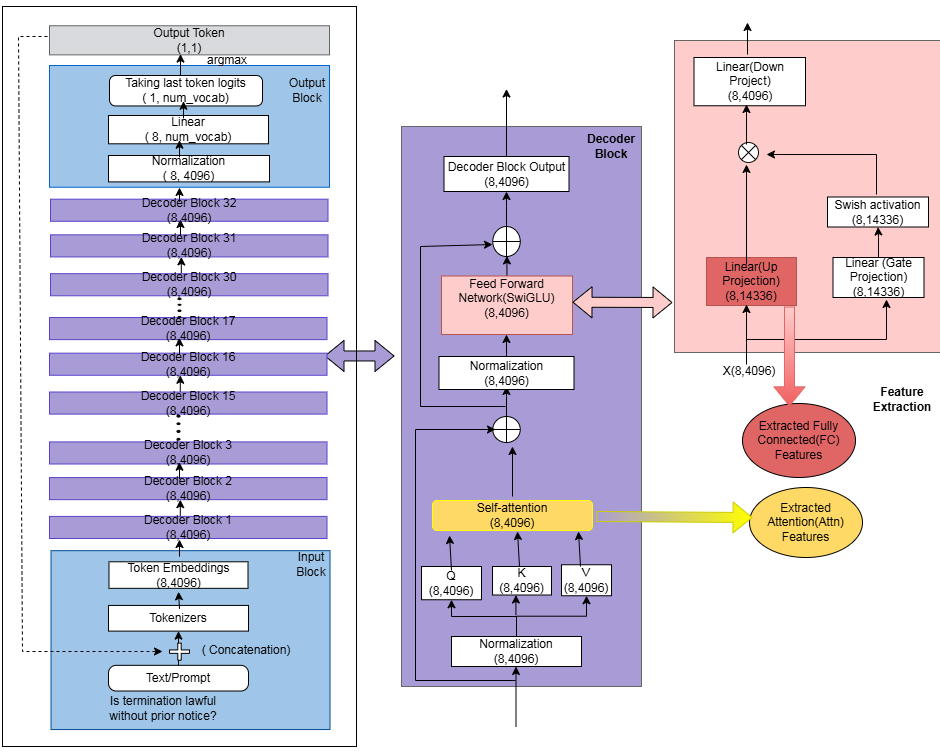}
\caption{Illustration of the architecture of an LLM (\textit{Llama-8b} with 32 decoder layers). The left side shows all layers; the central part shows the internal details of one decoder block. The internal artifacts used in our method are extracted from the self-attention module, and the Feed-Forward Network module. The numbers shown in each block (e.g., $(8, 4096)$, $(8, 14336)$) denote the per-token representation sizes, where the first value corresponds to the sequence length (number of tokens) and the second value corresponds to the embedding dimension of each token in that block. The term $\textit{num\_vocab}$ (in the output block) denotes the size of the model's vocabulary, i.e., the number of possible output tokens.} 
\label{fig:internal_features}
\end{figure}

Decoder-only LLMs process input text through multiple \textit{decoder layers}, where each layer produces intermediate representations that capture different levels of abstraction (e.g., lexical, syntactic, and semantic information). 
In this section, we describe the internal artifacts of LLMs that we use in our methodology for correctness detection on the LLM responses. 


A detailed illustration of the architecture of an LLM is presented in Figure~\ref{fig:internal_features}, highlighting the input processing, stacked decoder blocks, and the extraction of intermediate attention and feed-forward features used in our approach. The figure is based on LLaMA 3 models~\cite{grattafiori2024llama}.
In this figure, we take the example of the popular \textit{Llama-8b} LLM which has 32 decoder layers (shown in Figure~\ref{fig:internal_features} as `Decoder Block 1', `Decoder Block 2', and so on). Each decoder block processes the input in two main steps, as illustrated in the central part of Figure~\ref{fig:internal_features} by showing the internal structure of one decoder block. 
First, a \textit{self-attention} module helps the model focus on important words in the input by capturing relationships between tokens.
Next, a \textit{feed-forward network (FFN)} refines these representations through simple transformations. As detailed in the right-hand side of Figure~\ref{fig:internal_features}, the FFN block consists of two parallel linear projections: an up-projection and a gate projection. The gate-projected features are passed through a non-linear activation (Swish), and are then combined with the up projection via element-wise multiplication. The resulting representation is finally passed through a down-projection to return to the original dimension.
Residual connections are used around the Self-attention and FFN sub-layers to preserve information across layers.


Mathematically, Let $M$ denote an LLM with $L$ decoder layers, and $d_i$ denote an input. Now, let $M$ generate a response $r_i = (t_1, t_2, \dots, t_T)$, where $t_1$ denotes the first generated token, $t_2$ denotes the second generated token, and so on. 
For each decoder layer $l \in \{1, 2, \dots, L/2, \dots L\}$, we extract two types of internal features in our method for detecting correctness of responses: 
(i)~\textit{Fully Connected (FC) features (from the FFN)}, and 
(ii)~\textit{Self-Attention (Attn) features} (from the Self-attention block) corresponding to the first generated token $t_1$ as illustrated in Figure~\ref{fig:internal_features}.


\begin{table}[t]
\centering
\small
\setlength{\tabcolsep}{6pt}
\begin{tabular}{lll}
\toprule
\textbf{CD Scheme} & \textbf{LLM Layers used} & \textbf{Feature ($\phi_i$) computation} \\
\midrule
FC-Mid   & $L/2$ & $f^{(L/2)}(t_1)$ \\
FC-Last  & $L$   & $f^{(L)}(t_1)$ \\
Attn-Mid & $L/2$ & $a^{(L/2)}(t_1)$ \\
Attn-Last& $L$   & $a^{(L)}(t_1)$ \\
\midrule
FC-Mid3  & $L/2-1,\, L/2,\, L/2+1$ 
& $\frac{1}{3}\Big(f^{(L/2-1)}(t_1) + f^{(L/2)}(t_1) + f^{(L/2+1)}(t_1)\Big)$ \\
FC-Last3 & $L-2,\, L-1,\, L$ 
& $\frac{1}{3}\Big(f^{(L-2)}(t_1) + f^{(L-1)}(t_1) + f^{(L)}(t_1)\Big)$ \\
Attn-Mid3 & $L/2-1,\, L/2,\, L/2+1$ 
& $\frac{1}{3}\Big(a^{(L/2-1)}(t_1) + a^{(L/2)}(t_1) + a^{(L/2+1)}(t_1)\Big)$ \\
Attn-Last3 & $L-2,\, L-1,\, L$ 
& $\frac{1}{3}\Big(a^{(L-2)}(t_1) + a^{(L-1)}(t_1) + a^{(L)}(t_1)\Big)$ \\
\bottomrule
\end{tabular}
\caption{Feature sets $\phi_i$ (for input sample $d_i$) used in different Correctness Detection (CD) schemes. $L$ denotes the total number of decoder layers in the LLM. $f^{(l)}(t_1)$ and $a^{(l)}(t_1)$ respectively denote the \textit{Fully Connected} and \textit{Self-Attention} features for the first token at layer $l$, as explained in the text.}
\label{tab:CD-features}
\end{table}

Let $f^{(l)}(t_1)$ and $a^{(l)}(t_1)$ denote the \textit{FC} and \textit{Attn} features for the first token at layer $l$, respectively. 
We construct different feature sets $\phi_i$ (for the input $d_i$) by selecting and aggregating features from different parts of the network, as summarized in Table~\ref{tab:CD-features}. 
The first four rows indicate direct selection of representations from either the middle decoder layer ($L/2$) or the last decoder layer ($L$). 
Furthermore, prior work~\cite{yu2025context, song2025demystifying} has shown that aggregating representations across adjacent layers yields more stable features. 
Thus, we additionally construct feature sets from the \textit{middle three layers} (Mid3) and the \textit{last three layers} (Last3) by averaging the features across these layers; these are shown in the last four rows of Table~\ref{tab:CD-features}.

For example, in Figure~\ref{fig:internal_features}, which shows a 32-layer model ($L = 32$), the middle layer is $L/2 = 16$ and the last layer is $L = 32$. Thus, the Mid and Last settings use features from layers $16$ and $32$, respectively. In contrast, the Mid3 setting uses layers $\{15, 16, 17\}$ and the Last3 setting uses layers $\{30, 31, 32\}$, where features from these layers are averaged, as shown in Table~\ref{tab:CD-features}, to obtain more stable representations.


\section{Methodology}
\label{sec:methodology}



\begin{figure}[t]
\centering
\includegraphics[width=0.98\linewidth, height=9cm]{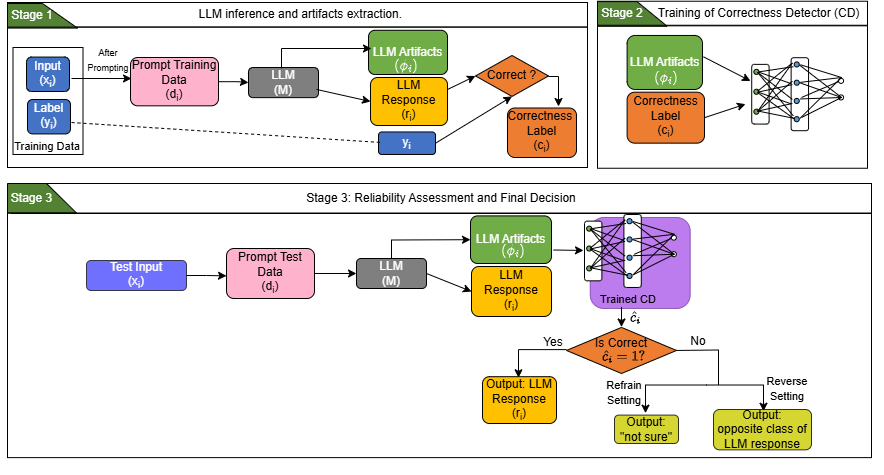}
\caption{Overview of the proposed three-stage framework: (1) LLM inference and artifact extraction, (2) Training of Correctness Detector (CD), and (3) Reliability assessment and final decision.}
\label{fig:method_overview}
\end{figure}

Our methodology consists of three main stages, as illustrated in Figure~\ref{fig:method_overview}: \textbf{(1) LLM Inference and Artifacts Extraction}, where internal artifacts are obtained from intermediate layers; \textbf{(2) Training of Correctness Detector (CD)}, where a lightweight artificial neural network is trained over these representations, to classify between correct and incorrect outputs; \textbf{(3) Reliability Assessment and Final Decision}, where the trained CD is used to assess the reliability of the LLM responses. A detailed description of each stage is provided below.

\paragraph{Dataset Splits:} 
We assume that our dataset
has two splits: the \textit{train} split $D_{\text{train}}$ and the \textit{test} split $D_{\text{test}}$. 
Here $D_{\text{train}} = \{(x_i, y_i)\}_{i=1}^{n_{\text{train}}}$  where $n_{\text{train}}$ is the number of samples in the train set, $x_i$ is an input document (e.g., a case facts), and  $y_i$ is the corresponding ground-truth label (e.g., `violation' or `no-violation')]. 
Similarly, $D_{\text{test}} = \{(x_i, y_i)\}_{i=1}^{n_{\text{test}}}$  is the test set, where $n_{\text{test}}$ is the number of samples in the test set.

\paragraph{Stage 1: LLM Inference and Artifacts Extraction.} 
In this stage, we use the \textit{train} set ($D_{\text{train}}$) where we know the correct label for every instance. 
We first use the LLM to generate a response for each instance $x_i$ in $D_{\text{train}}$, then extract the LLM's response and its internal artifacts (according to one of the approaches described in Section~\ref{sec:background}).
These extracted artifacts are later used as features to train the CD.

Since LLMs operate on prompted inputs, we construct an input $d_i = \mathcal{P}(x_i)$, using a predefined prompt template  $\mathcal{P}(\cdot)$ (described in Section~\ref{subsec:prompts}). For each input $d_i$, the LLM ($M$) generates a response $r_i$.
Feature vectors ($\phi_i$) are then extracted from its internal artifacts at the first generated token of $r_i$, as described in Section~\ref{sec:background}. We define the correctness label $c_i \in \{0,1\}$ such that a response is \textit{Incorrect} ($c_i = 0$) if the LLM-generated response $r_i$ does not match the ground-truth label $y_i$, and \textit{Correct} ($c_i = 1$) otherwise. Finally, in this stage, we obtain $\{(\phi_i, c_i)\}_{i=1}^{n_{\text{train}}}$ from $D_{\text{train}}$, where $\phi_i$ denotes the feature vector extracted from the LLM, and $c_i \in \{0,1\}$ is the correctness label, both for the instance $x_i$.

\paragraph{Stage 2: Training of Correctness Detector (CD).} 
Here, we train a binary classifier to predict whether an LLM-generated response is correct, based on its internal artifacts. 
So, our objective is to train a binary CD, $C: \phi_i \rightarrow \{0,1\}$, where $C$ predicts the correctness label $c_i$. 
For the training and evaluation of CD, we  split $\{(\phi_i, c_i)\}_{i=1}^{n_{\text{train}}}$ (obtained from $D_{\text{train}}$ in Stage~1) in 60:20:20 ratio into \textit{CD-train}, \textit{CD-val}, and \textit{CD-test} subsets. 
The CD is trained using \textit{CD-train} and \textit{CD-val}, and its performance is evaluated over \textit{CD-test}.
 
Following prior works~\cite{datta2026lowresource, snyder2023early}, our CD is a lightweight feed-forward single-layer neural network with a hidden dimension of 256. We train separate CD models on each type of $\phi_i$ shown in Table~\ref{tab:CD-features} independently, resulting in multiple CDs for different types: \textit{FC-Mid3}, \textit{FC-Last3}, \textit{Attn-Mid3}, and \textit{Attn-Last3}. 
For instance, as explained in Section~\ref{sec:background}, \textit{FC-Mid3} denotes the mean aggregation of fully connected layer representations from the middle three layers of the LLM, extracted at the first generated token of the LLM response. This enables a systematic comparison of which internal representations are most effective for correctness detection.

\paragraph{Stage 3: Reliability Assessment and Final Decision.} 
In this stage, we utilize our trained CD to assess the reliability of the LLM. We consider the overall system to be LLM+CD where the response by the LLM is verified by the CD, and modified if deemed necessary. 

We use the test set  $D_{\text{test}}$ for this stage.
Given an input $d_i \in D_{\text{test}}$, we first feed $d_i$ into the LLM and obtain the LLM response $r_i$ and the features $\phi_i$ from the internal artifacts of the LLM. We then use the trained CD (that was trained in Stage~2) to predict  class label $\hat{c}_i$ where $\hat{c}_i = 1$ indicates that the LLM response is predicted as correct by the CD, and $\hat{c}_i = 0$ indicates that it is predicted as incorrect by the CD. Based on this prediction, we consider two settings:

\noindent \textbf{(i) \textsc{Refrain} Setting:} 
If $\hat{c}_i = 0$, the LLM+CD system refrains from generating the response 
and returns a fallback output (\textit{``not sure''}). 
If $\hat{c}_i = 1$, 
the original response generated by the LLM is returned.
This setting is motivated by the principle that, in high-stakes scenarios such as law, it is preferable for a system to abstain/refrain from answering rather than provide a potentially incorrect answer. In other words, expressing uncertainty (e.g., \textit{``not sure''}) is often safer than returning a wrong prediction. For instance, if the system refrains from answering, the question can be posed to a human Law expert. 

\noindent \textbf{(ii) \textsc{Reverse} Setting:}  
Alternatively, instead of refraining, we leverage the CD prediction as an uncertainty signal. If $\hat{c}_i = 0$, we interpret this as an indication that the LLM prediction is likely incorrect according to the CD and replace it with the opposite class. If $\hat{c}_i = 1$ (i.e., LLM prediction is correct according to the CD), then the original LLM response is retained. The primary motivation for this setting is to explore whether the CD's predictions can be trusted more than the LLM's prediction. 

Note that the \textsc{Reverse} setting is suitable only for binary classification. For more complex tasks, such as classification with more than two classes, different variants of this setting can be explored, such as asking the LLM to predict again after adding more details to the prompt; we leave exploring these variants for future work.

\paragraph{Different Schemes in Our Final Systems.} We develop a final system by incorporating a trained CD with the LLM; we refer to this final system as \textbf{LLM+CD}.   
Overall, we consider nine systems -- one \textit{LLM-only} baseline system, and four schemes each for \textsc{Refrain} and \textsc{Reverse} settings: LLM+\{\textit{FC-Mid3} / \textit{FC-Last3} / \textit{Attn-Mid3} / \textit{Attn-Last3}\}-\{RFN / REV\}.
The \textit{LLM-only} scheme refers to an off-the-shelf LLM without integrating any CD. 
There is one LLM+CD system corresponding to every row of Table~\ref{tab:CD-features}.
For instance, the \textit{LLM+FC-Mid3-RFN} scheme integrates the CD trained on \textit{FC-Mid3} features, and considers the \textsc{Refrain} setting. 
Similarly, the \textit{LLM+FC-Mid3-REV} scheme applies the same trained CD, considering the \textsc{Reverse} setting.

\paragraph{Implementations of the methods.} The implementations of the proposed methods are publicly available at: \url{https://github.com/iamDebtanu/LLM_Correctness_Detection_Legal}.

\section{Experimental Setup}
\label{sec:experiments}

This section describes in detail the datasets / tasks, LLMs and evaluation metrics used in our experiments.

\subsection{Datasets}
\label{subsec:datasets}
In this section, we describe the datasets that we used to evaluate our approaches. We consider two popular datasets for binary classification task in the legal domain, covering different jurisdictions: 
(i)~ECHR dataset for statute violation prediction~\cite{Chalkidis1}, and (ii) ILDC dataset for bail decision prediction~\cite{Vijit1}. The details of these datasets are described below:

$\bullet$ \textbf{Statute violation prediction with ECHR Dataset.}
The ECHR dataset~\cite{Chalkidis1}  consists of cases from the European Court of Human Rights, where the task is to predict whether a human rights violation has occurred based on the provided case. Each case is associated with articles of the European Convention on Human Rights, and the task is typically framed as predicting violations for specific articles based on the case facts. The dataset contains long textual case documents describing the facts, arguments, and judicial reasoning, making it suitable for evaluating the legal reasoning capabilities of models.
For this work, we use the classification setting as violation (1) vs. no-violation (0). Table~\ref{tab:echr_full} summarizes the dataset statistics and class distribution in the ECHR dataset which is split into the train and test splits.

\begin{table}[tb]
\centering
\caption{Dataset statistics and class distribution of the ECHR dataset.}
\label{tab:echr_full}
\begin{tabular}{|l|c|c|c|}
\hline
\textbf{Split} & \textbf{Samples} & \textbf{No-Violation (0)} & \textbf{Violation (1)} \\
\hline
Train ($D_{\text{train}}$) & 7,100 & 3,549 & 3,551 \\
Test ($D_{\text{test}}$) & 2,998 & 1,024 & 1,974 \\
\hline
\end{tabular}
\end{table}

$\bullet$ \textbf{Bail decision prediction with ILDC Dataset.}
The ILDC dataset~\cite{Vijit1} is a popular bail decision prediction dataset in the Indian Judiciary. It consists of bail appeal cases from the Supreme Court of India, annotated with a final verdict of appeal rejection or acceptance. Here, we consider the binary classification setting, where cases (bail applications) are labeled as accepted (1) or rejected (0). Specifically, the task is to predict the outcome of a case based on its facts and legal issues. We use the $\textit{ILDC\textsubscript{Single}}$ split, which contains cases with a single petition and a corresponding single decision. The dataset statistics and class distribution are reported in Table~\ref{tab:ildc_stats}.

\begin{table}[tb]
\centering
\caption{Dataset statistics and class distribution of the ILDC dataset}
\label{tab:ildc_stats}
\begin{tabular}{|l|c|c|c|}
\hline
\textbf{Split} & \textbf{Samples} & \textbf{Reject (0)} & \textbf{Accept (1)} \\
\hline
Train ($D_{\text{train}}$) & 5,082 & 3,147 & 1,935 \\
Test ($D_{\text{test}}$) & 1,517 & 755   & 762   \\
\hline
\end{tabular}
\end{table}

\subsection{LLMs and Prompts}
\label{subsec:prompts}

We conduct comprehensive experiments using three open-source instruction-tuned LLMs to assess our approaches, namely Llama-3.1-8B-Instruct\footnote{\url{https://huggingface.co/meta-llama/Llama-3.1-8B-Instruct}} (Llama-8b in short), Mistral-7b-instruct-v0.3\footnote{\url{https://huggingface.co/mistralai/Mistral-7B-Instruct-v0.3}} (Mistral-7b), and Qwen2.5-7B-Instruct\footnote{\url{https://huggingface.co/Qwen/Qwen2.5-7B-Instruct}} (Qwen-7b in short).


We design a \textit{system-user}-based prompt template to guide these LLMs. The prompts for each task/dataset consist of a \textit{system} instruction that defines the legal context and task objective, followed by a \textit{user} query / instruction containing the case description. The prompts used for the ECHR and ILDC datasets are shown in Figure~\ref{fig:prompt_echr} and Figure~\ref{fig:prompt_ildc}, respectively.

\begin{figure}[tb]
    \centering
    \begin{tcolorbox}[colback=gray!15, colframe=black, boxrule=1pt, width=0.95\textwidth, fontupper=\normalfont\small, boxsep=1pt, left=2pt, right=2pt, top=1pt, bottom=1pt, halign=justify]
    \footnotesize
    \textbf{System Prompt:}
    You are a legal expert specialized in human rights law under the European Convention on Human Rights (ECHR). Your task is to analyze case proceedings from the European Court of Human Rights and determine whether the facts disclose a violation of a human rights article under the ECHR. Your decision must be strictly based on the facts provided and the provisions of the European Convention on Human Rights. Output only one word: either ``Violation'' or ``No-Violation''.

    \textbf{User Prompt:}
    Based on the European Convention on Human Rights, determine whether the following case proceeding shows a violation of a human rights article.

    \textbf{Case Proceeding:} $\langle X \rangle$

    \textbf{Decision:}
    \end{tcolorbox}
    \captionof{figure}{Prompt template used for ECHR Dataset for statute violation prediction task.}
    \label{fig:prompt_echr}
\end{figure}


\begin{figure}[tb]
    \centering
    \begin{tcolorbox}[colback=gray!15, colframe=black, boxrule=1pt, width=0.95\textwidth, fontupper=\normalfont\small, boxsep=1pt, left=2pt, right=2pt, top=1pt, bottom=1pt, halign=justify]
    \footnotesize

    \textbf{System Prompt:}
    You are a legal expert specialized in Indian law. Your task is to analyze the given case proceeding strictly based on principles of Indian law and established judicial reasoning, and determine whether the appeal should be accepted or rejected. Your decision must be solely based on the facts and legal issues presented in the case. Output only one word: either ``Accept'' or ``Reject''.

    \textbf{User Prompt:} Predict the outcome of the appeal in the following case proceeding based strictly on Indian law. 
    
    \textbf{Case Proceeding:} $\langle X \rangle$
    
    \textbf{Decision:}
    \end{tcolorbox}
    \captionof{figure}{Prompt template used for ILDC Dataset for bail decision prediction task.}
    \label{fig:prompt_ildc}
\end{figure}

\subsection{Infrastructure and Hyperparameters}
All experiments are conducted on a server with 2 x NVIDIA A6000 GPUs, each having 48 GB of GPU memory. We summarize the hyperparameter settings for both LLM inference and CD training below: 

$\bullet$ \textbf{LLM inference:} 
Decoding Strategy: Greedy (temperature = 0); 
Max Generation Length: 5 tokens;
Stopping Criterion: EOS token or Max Generation Length.

$\bullet$ \textbf{CD Training:} Activation: ReLU; 
Dropout: 0.1;
Optimizer: AdamW;
Learning Rate: $1 \times 10^{-4}$;
Batch Size: 128;
Loss Function: Cross-Entropy;
Early Stopping: Patience = 5 (based on validation ROC-AUC).\\
All settings are kept uniform across models and datasets to ensure fair comparison.

\subsection{Evaluation Metrics}

\paragraph{Metrics for Evaluating Correctness Detector (CD).} 
Recall from Section~\ref{sec:methodology} that we held-out a 20\% subset of the samples in the training set for evaluation of the CD. This subset, referred to as \textit{CD-test}, is used for evaluation of the CD. 
To evaluate the performance of CD, we report the AUROC (Area Under the Receiver Operating Characteristic)~\cite{fawcett2006introduction,hanley1982meaning} metric due to its robustness to class imbalance and threshold independence. Since the proportion of correct and incorrect responses is heavily skewed, standard accuracy may provide a misleading estimate of performance; hence AUROC is usually reported in most studies on detecting correctness of LLM responses~\cite{snyder2023early, datta2026lowresource}. 
AUROC ranges from 0 to 1, where 1 indicates perfect classification, 0.5 corresponds to random guessing, and values below 0.5 indicate worse-than-random performance. For ease of interpretation, we report AUROC scaled by 100.









\paragraph{Metrics for Evaluating the Performance of LLM-only and LLM+CD Systems.} 
We design five metrics to evaluate the LLM-only and LLM+CD systems (under two settings \textsc{Refrain} and \textsc{Reverse}, as described in Section~\ref{sec:methodology}): \textbf{Coverage}, \textbf{Accuracy}, \textbf{CD-Gain}, and \textbf{CD-Cost}.


\textbf{$\bullet$ Coverage:} This measures the proportion of instances for which the system outputs a valid label for the task.
   We consider the set of all possible valid labels (`\textit{Accept}' or `\textit{Reject}' for ILDC, and `\textit{Violation}' or `\textit{No-Violation}' for ECHR) which the LLMs are specifically asked to output (see the prompts in Fig.~\ref{fig:prompt_echr} and Fig.~\ref{fig:prompt_ildc}). 
   For each sample in the \textit{test} set ($D_{\text{test}}$), we look into the first few (at most five) tokens in the LLM-generated response and check whether any of the valid labels is present.
   The coverage can be less than 100\% in two scenarios: 
   (i)~when a valid label is not present in the first few tokens generated by the LLM, and 
   (ii)~in \textsc{Refrain} setting, when CD predicts that the LLM response is incorrect and consequently, the LLM+CD system refrains from producing the response and returns a fallback output (``not sure"). 

\[
\text{Coverage} = 
\frac{\mbox{\# instances for which system outputs a valid label}}{\mbox{\# total instances}} \times 100
\]

The rest of the metrics only consider those instances where a system outputs a valid label.

\textbf{$\bullet$  Accuracy:} This measures the proportion of correct label predictions among the valid answered instances, reflecting the system's reliability.

\[
\text{Accuracy} = \frac{\mbox{\# instances for which system outputs the correct label}}{\mbox{\# instances for which system outputs a valid label}} \times 100
\]

~\\
\noindent \textit{The next three metrics are applicable only to the LLM+CD systems, and specifically capture the trade-off of adding the CD module to the LLM} -- while some wrong answers by the LLM can be refrained from / corrected (gain), some originally correct answers by the LLM may also be wrongly refrained from / degraded (Cost) due to addition of the CD.

\textbf{$\bullet$  CD-Gain: }This measures the fraction of incorrect valid LLM predictions/responses that are either corrected (in \textsc{Reverse} setting) or  refrained (in \textsc{Refrain} setting) after integrating CD.
\[
\text{CD-Gain} =
\frac{\mbox{\# instances where incorrect valid LLM response is corrected or refrained}}
{\mbox{\# instances for which LLM outputs incorrect valid response}} \times 100
\]

\textbf{$\bullet$  CD-Cost: }This measures the fraction of correct valid LLM responses/predictions that are degraded (made wrong) or unnecessarily refrained by CD.
\[
\text{CD-Cost} =
\frac{\mbox{\# instances where correct LLM responses are refrained or made wrong}}
{\mbox{\# instances for which LLM outputs correct valid response}} \times 100
\]

\textbf{$\bullet$ Net-Gain: }This measures the overall net improvement achieved by adding CD, by balancing the corrected/refrained incorrect LLM responses against the correct LLM responses that are degraded or unnecessarily refrained.


\[
\text{\footnotesize Net-Gain =}
\frac{\footnotesize
\begin{array}{c}
\text{\#incorrect valid LLM responses corrected/refrained} 
- 
\text{\#correct LLM responses refrained/made wrong}
\end{array}
}
{\text{\footnotesize \# instances for which LLM outputs a valid response}}
\times \text{\footnotesize 100}
\]

\section{Results and Observations}


\begin{table}[t]
\centering
\caption{AUROC scores of different Correctness Detectors across models over the \textit{CD-test} split of both ILDC and ECHR datasets. Best results in each LLM are highlighted in \textbf{bold}.}
\label{tab:CD}
\begin{tabular}{|c|ccc|ccc|}
\hline
\multirow{3}{*}{CD} & \multicolumn{3}{c|}{ILDC-\textit{CD-test}} & \multicolumn{3}{c|}{ECHR-\textit{CD-test}} \\ \cline{2-7}
 & Llama-8b & Mistral-7b & Qwen-7b & Llama-8b & Mistral-7b & Qwen-7b \\
\hline
\hline
FC-Mid   & 93.94 & 84.24 & 64.88 & 90.55 & 87.82 & 72.42 \\
FC-Last  & 94.72 & 93.39 & 95.31 & 90.52 & 90.44 & 83.57 \\
FC-Mid3  & 92.06 & 89.33 & 75.34 & 90.54 & 89.95 & 77.16 \\
FC-Last3 & \textbf{95.12} & \textbf{93.40} & 95.18 & 91.19 & \textbf{90.79} & 85.32 \\
\hline
Attn-Mid  & 91.22 & 85.62 & 70.73 & 89.80 & 87.40 & 72.75 \\
Attn-Last & 92.83 & 91.80 & 92.88 & 90.40 & 87.75 & \textbf{88.71} \\
Attn-Mid3 & 94.17 & 90.09 & 75.62 & 90.80 & 87.21 & 79.75 \\
Attn-Last3& 92.23 & 92.85 & \textbf{95.36} & \textbf{91.89} & 88.95 & 88.27 \\
\hline
\end{tabular}
\end{table}

\paragraph{Task Performance of LLMs.}
We first evaluate the off-the-shelf performance of the LLMs on the \textit{test} splits of the ILDC and ECHR datasets. As shown in Table~\ref{tab:main} (LLM-only rows), all models achieve moderate performance for both tasks, with accuracy ranging between approximately 54\% and 61\%. \textit{These results demonstrate the inherent difficulty of these tasks in the legal domain, highlighting the need for a reliability mechanism such as a correctness detector, which aims to develop a reliable system.}

\paragraph{Performances of Correctness Detectors.}
Table~\ref{tab:CD} reports the AUROC scores of different CD variants over the \textit{CD-test} split of both ILDC and ECHR datasets. 
Overall, detectors showcase strong performance, with most settings exceeding an AUROC score of 85, indicating effective discrimination between correct and incorrect LLM responses. \textbf{We observe that multi-layer representations (\textit{Mid3} and \textit{Last3}) consistently outperform or remain competitive with single-layer variants (\textit{Mid} and \textit{Last}), suggesting that correctness signals are distributed across layers.} While both fully connected (FC) and attention-based (Attn) features perform well, their relative strengths vary across settings. Although \textit{Last3} variants mostly achieve the higher AUROC, \textit{Mid3} variants remain competitive. Based on the consistently strong and stable performance, we select detectors with these four settings -- \textit{FC-Mid3}, \textit{FC-Last3}, \textit{Attn-Mid3}, and \textit{Attn-Last3} -- for subsequent evaluation of the LLM+CD systems.

\begin{table*}[t]
\centering
\footnotesize
\caption{
Performance of LLM+CD schemes under \textsc{Refrain} (RFN) and \textsc{Reverse} (REV) settings across models over the test split of ILDC and ECHR. For each LLM, the best value in each metric is highlighted in \textbf{bold}.
}
\label{tab:main}
\setlength{\tabcolsep}{2pt}

\begin{tabular}{|l|ccccc|ccccc|}
\hline

\multirow{2}{*}{LLM+CD Scheme} 
& \multicolumn{5}{c|}{\textbf{ILDC Dataset}} 
& \multicolumn{5}{c|}{\textbf{ECHR Dataset}} \\
\cline{2-6} \cline{7-11}
& Accuracy & Coverage & CD-Gain & CD-Cost & Net-Gain 
& Accuracy & Coverage & CD-Gain & CD-Cost & Net-Gain \\
\hline
\hline

\multicolumn{11}{|c|}{\textbf{Llama-8b}} \\
\hline
LLM-only & 54.84 & \textbf{100} & - & - & - & 55.17 & \textbf{98.36} & - & - & - \\
FC-Mid3-RFN    & 78.06 & 50.50    & 76.06 & 30.13 & 18.06  & \textbf{79.39} & 45.83    & \textbf{78.59} & 33.24 & 17.02 \\
FC-Mid3-REV    & 72.85    & \textbf{100}   & 76.06 & 30.13 & 18.06   & 71.63    & \textbf{98.36}    & \textbf{78.59}    & 33.24 & 17.02\\
FC-Last3-RFN    & \textbf{84.52} & 43.44 & \textbf{85.25} & 34.14 & \textbf{19.91}    & 71.40  & 55.95    & 63.77   & 26.30 & 14.04 \\
FC-Last3-REV    & 74.85    & \textbf{100}   & \textbf{85.25} & 34.14 & \textbf{19.91}   & 69.20  & \textbf{98.36} & 63.77    & 26.30 & 14.04\\
Attn-Mid3-RFN  & 78.88 & 51.81 & 76.06 & \textbf{27.04} & 19.51 & 71.40  & 55.95    & 63.77    & 26.30 & 14.04 \\
Attn-Mid3-REV  & 74.35 & \textbf{100}   & 76.06 & \textbf{27.04} & 19.51 &69.20  & \textbf{98.36} & 63.77    & 26.30 & 14.04\\
Attn-Last3-RFN  & 78.73 & 40.60  & 81.17 & 43.05 &13.12  & 74.26    & 61.94    & 63.09    & \textbf{16.23} & \textbf{19.53}\\
Attn-Last3-REV  & 68.07    & \textbf{100}   & 81.17 & 43.05 &13.12   & 75.14    & \textbf{98.36} & 63.09    & \textbf{16.23} & \textbf{19.53} \\

\hline
\multicolumn{11}{|c|}{\textbf{Mistral-7b}} \\
\hline

LLM-only      & 58.78 & \textbf{99.14} & - & - & - & 61.27 & \textbf{99.90} & - & - & - \\
FC-Mid3-RFN    & 74.87 & 65.59 & 61.61 & \textbf{19.11} & \textbf{14.19} & 81.16   & 50.63    & 75.09 & 33.13 & 8.85 \\
FC-Mid3-REV    & 73.15 & \textbf{99.14} & 61.61  & \textbf{19.11} & \textbf{14.19} & 70.14   & \textbf{99.90}   & 75.09    & 33.13 & 8.85 \\
FC-Last3-RFN    & 77.29 & 54.06    & 71.00    & 30.30 & 11.45 & \textbf{82.15}   & 52.85    & 74.66    & 30.04 & 10.69\\
FC-Last3-REV    & 70.30  & \textbf{99.14} & 71.00  & 30.30 & 11.45  & 72.26   & \textbf{99.90}   & 74.66  & 30.04 & 10.69 \\
Attn-Mid3-RFN  & \textbf{79.15} & 54.38 & \textbf{73.06}  & 28.13 & 13.82  & 78.06 & 60.37 & 65.78    & \textbf{23.13} & \textbf{11.39} \\
Attn-Mid3-REV  & 72.64 & \textbf{99.14} & \textbf{73.06}  & 28.13 &13.82   & 73.07    & \textbf{99.90}   & 65.78    & \textbf{23.13} & \textbf{11.39}\\
Attn-Last3-RFN  & 71.80  & 58.67 & 61.13    & 30.30 & 7.41 & 81.50  & 48.78    & \textbf{76.55}    & 35.27 & 8.18\\
Attn-Last3-REV  & 66.35 & \textbf{99.14} & 61.13    & 30.30 & 7.41 & 69.44 & \textbf{99.90}  & \textbf{76.55}  & 35.27 & 8.18\\

\hline
\multicolumn{11}{|c|}{\textbf{Qwen-7b}} \\
\hline

LLM-only        & 59.20 & \textbf{100} & - & - & - & 54.14 & \textbf{100} & - & - & - \\
FC-Mid3-RFN    & 60.17    & 95.72    & 5.58  & \textbf{3.33} & 0.33 & 74.16   & 33.69    & \textbf{79.10} & 57.17 & 0.47\\
FC-Mid3-REV    & 59.86    & \textbf{100}   & 5.58  & \textbf{3.33} & 0.33  & 58.20 & \textbf{100}   & \textbf{79.10} & 57.17 & 0.47\\
FC-Last3-RFN    & 74.17    & 72.38    & 54.05   & 9.52  & \textbf{16.43} & 66.16   & 70.15    & 43.40 & \textbf{20.05} & 6.84\\
FC-Last3-REV    & \textbf{75.74}    & \textbf{100}   & 54.05   & 9.52 & \textbf{16.43} & 65.23   & \textbf{100}   & 43.40 & \textbf{20.05} & 6.84 \\
Attn-Mid3-RFN  & 65.64    & 68.16    & 44.56   & 24.26  & 3.75 & 65.67   & 70.15    & 43.40 & 20.12 & 6.85 \\
Attn-Mid3-REV  & 63.16    & \textbf{100}   & 44.56   & 24.26 & 3.75  & 65.16   & \textbf{100}   & 43.40 & 20.12 & 6.85\\
Attn-Last3-RFN  & 73.33 & 68.49    & \textbf{55.25} & 15.12 & 13.58 & \textbf{75.33} & 60.87    & 64.75   & 21.12 & \textbf{15.31} \\
Attn-Last3-REV  & 72.78    & \textbf{100}   & \textbf{55.25} & 15.12 &13.58 & 73.05   & \textbf{100}   & 64.75   & 21.12 & \textbf{15.31}\\

\hline
\end{tabular}
\end{table*}

\paragraph{Performance of LLM+CD Framework.}
Table~\ref{tab:main} presents the performance of our LLM+CD systems under both \textsc{Refrain} (RFN) and \textsc{Reverse} (REV) settings over the \textit{test} splits of the ILDC and ECHR datasets.

$\bullet$ \textbf{Effect of Correctness Detection.}
Incorporating CD consistently improves performance across all models. For instance, the accuracy for Llama-8b improves from $\sim$55\% (LLM-only scheme) to over 84\% in the LLM+FC-Last3-RFN scheme. Similar trends are observed for Mistral-7b and Qwen-7b as well, \textit{indicating that internal representations provide strong signals for detecting incorrect predictions}.

$\bullet$ \textbf{{\textsc{Refrain} vs \textsc{Reverse} Trade-off.}}
The schemes with \textsc{Refrain} settings achieve higher accuracy by refraining from uncertain predictions, at the cost of reduced coverage. In contrast, the \textsc{Reverse} setting maintains near full coverage ($100\%$) by handling cases where the model refrains from answering, though with slightly lower accuracy; however, in some cases, coverage falls a bit from 100\% when the LLM response lies outside the valid label space. 

$\bullet$ \textbf{Observations from CD-Aware Metrics.}
The CD-Gain values are consistently high across models, showing that a large fraction of incorrect LLM responses are successfully corrected or safely handled. Meanwhile, CD-Cost remains moderate across schemes (except for FC-Mid3 in Qwen-7b), indicating that only a small portion of correct predictions are negatively affected. As a result, all schemes achieve positive Net-Gain values, showing that the overall improvement is larger than the errors introduced by the correctness detector. This \textit{highlights the effectiveness of the correctness detector in improving overall decision reliability}.


$\bullet$ \textbf{Generalization Across Datasets.}
The significant performance improvements from the LLM-only scheme to the LLM+CD system are consistent across models for both ILDC and ECHR datasets, \textit{demonstrating the robustness and general applicability of our proposed methodology}.

\section{Conclusion}

In this work, we develop approaches for early detection of incorrect LLM predictions in legal classification tasks by leveraging LLMs' internal parameters (features). 
We train a correctness detector (CD) using these features to effectively identify incorrect LLM responses. Our results on the bail prediction (using the ILDC dataset) and statute violation prediction (using ECHR) tasks demonstrate that \textit{an LLM+CD system (after integrating CD into LLMs) significantly improves decision reliability compared to the LLMs' off-the-shelf performances}. Importantly, this trend holds across LLMs. Further, the \textsc{Refrain} setting achieves higher accuracy by avoiding uncertain predictions, while the \textsc{Reverse} setting maintains full coverage with competitive performance. 
Moreover, high CD-Gain and moderate CD-Cost demonstrate that the LLM+CD system effectively corrects incorrect LLM responses with limited impact on correct ones. 
Overall, \textbf{our results highlight the potential of internal feature-based correctness detection for enhancing the reliability of LLMs in classification tasks in the legal domain}.

\section{Limitations and Future Work}
While our approach demonstrates strong performance improvements, it has some limitations that warrant further investigation. 
Our approach requires access to internal parameters during inference, which may not be feasible for closed-source or API-based models such as GPT-4 or GPT-5. 
Additionally, our evaluation in this paper is limited to binary classification tasks.
For more complex classification tasks (e.g., multi-class classification), different settings can be explored, such as asking the LLM to predict again after adding more details to the prompt; we leave exploring these variants for future work.


\bibliography{main_bib}

\end{document}